\documentclass{article}
\usepackage{iclr2027_conference,times}
\usepackage[T1]{fontenc}
\usepackage[utf8]{inputenc}
\newcommand{\lang}[2]{#1}
\usepackage{amsmath,amssymb,booktabs,graphicx,array,tabularx,multirow,xcolor,url}
\usepackage{microtype}
\usepackage{xr-hyper}
\usepackage{placeins,float}
\usepackage{hyperref}
\usepackage{cleveref}

\usepackage{xspace}
\definecolor{pendingcolor}{HTML}{C00000}

\newcommand{\sg}{\operatorname{sg}}
\newcommand{\softmax}{\operatorname{softmax}}
\newcommand{\risk}{R_{\mathrm{final}}}

\newcommand{\na}{\textsf{N/A}}

\newcommand{\assetpath}[1]{{\scriptsize\textcolor{gray}{\lang{Asset}{图像路径}: \nolinkurl{#1}}}}
\newcommand{\datapath}[1]{{\scriptsize\textcolor{gray}{\lang{Data slot}{数据接口}: \nolinkurl{#1}}}}
\newcommand{\mainref}[1]{\ref{#1}}

\hypersetup{colorlinks=true,linkcolor=blue!45!black,citecolor=blue!45!black,urlcolor=blue!45!black,pdftitle={VCRE-Fib -- View-Conditioned Regional Evidence},pdfauthor={Ziyang Xu, Shuli An, Hao Zhou, Haitian Zhong, Tingting Wu, Tao Wang, Kun Yang, Tieyong Zeng}}

\newcommand{\fitresulttable}[1]{#1}

\title{VCRE-Fib: View-Conditioned Regional\newline Evidence for Fine-Grained Ultrasound\newline Grading of \textnormal{\emph{Schistosoma japonicum}}-Associated\newline Liver Fibrosis\thanks{Code is available at: \url{https://github.com/StatXzy7/vcre-fib}}}
\hypersetup{pdftitle={VCRE-Fib: View-Conditioned Regional Evidence for Fine-Grained Ultrasound Grading of Schistosoma japonicum-Associated Liver Fibrosis}}
\author{%
Ziyang Xu$^{1}$, Shuli An$^{2}$, Hao Zhou$^{1}$, Haitian Zhong$^{3,4}$,\\
\textbf{Tingting Wu$^{5}$, Tao Wang$^{2}$, Kun Yang$^{2}$\thanks{Corresponding authors.}, Tieyong Zeng$^{6,7}$\footnotemark[2]}\\
$^{1}$Department of Mathematics, The Chinese University of Hong Kong\\
Hong Kong, China\\
$^{2}$Jiangsu Provincial Key Laboratory on Parasite and Vector Control Technology,\\
National Health Commission Key Laboratory of Parasitic Disease Control and Prevention,\\
Jiangsu Institute of Parasitic Diseases, Wuxi, Jiangsu, China\\
$^{3}$New Laboratory of Pattern Recognition (NLPR),\\
State Key Laboratory of Multimodal Artificial Intelligence Systems (MAIS),\\
Institute of Automation, Chinese Academy of Sciences\\
$^{4}$Zhongguancun Academy\\
$^{5}$School of Science, Nanjing University of Posts and Telecommunications\\
Nanjing, China\\
$^{6}$Institute for Advanced Study, Beijing Normal-Hong Kong Baptist University\\
Zhuhai, China\\
$^{7}$School of Mathematics and Statistics, Guangzhou Nanfang College\\
Guangzhou, China\\
\texttt{ziyang.xu@link.cuhk.edu.hk}\\
\texttt{yangkun@jipd.com, tieyongzeng@bnbu.edu.cn}
}
\iclrfinalcopy
\begin{document}
\maketitle
\lhead{Preprint}
\begin{abstract}
Accurate assessment of \emph{Schistosoma japonicum}-associated liver fibrosis is essential for disease management and long-term follow-up in endemic regions. Ultrasound provides non-invasive imaging, but complex local echogenic patterns and anatomical structures make fine-grained grading challenging. Existing deep learning methods can predict fibrosis scores, yet directly incorporating acquisition views and regional cues into grading while retaining spatial information for inspection remains an open problem. Here we present VCRE-Fib, a view-conditioned regional evidence framework that integrates anatomical context, local information, and global image assessment for fine-grained ultrasound grading. The framework forms view-conditioned local grading evidence before spatial pooling, uses weak localization to guide its aggregation, and combines it with global predictions. Image-only inference jointly returns a fibrosis score, acquisition view, and candidate abnormal-region map. We developed and evaluated the method on a re-curated cohort of 108,709 ultrasound images from 6,373 patients across 35 centers. On a patient-disjoint test set of 4,107 images from 240 patients across four centers, VCRE-Fib reduced the prespecified composite grading risk by 7.115\% relative to SFibAI trained and evaluated on the same data split. Image-level mean absolute error decreased from 0.391 to 0.378, alongside lower patient-max, patient-median, and center-balanced risks. The full model also achieved lower composite grading risk than variants that separately removed view conditioning or weak localization. These results support incorporating anatomical context and regional evidence into ultrasound grading while exposing spatial predictions for inspection alongside severity estimates.

\end{abstract}
\section{Introduction}\label{sec:introduction}
Liver fibrosis caused by \emph{Schistosoma japonicum} is a persistent concern in disease management in endemic regions and can be accompanied by portal hypertension and severe hepatosplenic complications \citep{colley2014schistosomiasis,mcmanus2018schistosomiasis}. In endemic settings, including China, assessment of established hepatic morbidity complements infection control \citep{mcmanus2010china,lo2022who}. Longitudinal evidence shows that severe parenchymal fibrosis can persist for years despite antiparasitic treatment \citep{carlton2010morbidity}. Repeatable imaging assessment of disease severity therefore provides an important foundation for population screening, assessment of morbidity, and long-term follow-up.

B-mode ultrasound non-invasively depicts hepatic parenchymal and portal abnormalities, providing an imaging tool suitable for repeated assessment. However, the ultrasound appearances of \emph{S. japonicum}-associated fibrosis have complex spatial patterns: interseptal fibrosis can produce network-like or fish-scale-like echoes, whereas portal abnormalities have a different anatomical distribution. The Basel ultrasonography protocol emphasizes distinguishing these patterns \citep{richter2025basel}. These features make local echoes, structural distribution, and anatomical context relevant to severity assessment. For automated grading, this raises two linked requirements: extracting information that supports fine-grained scoring from complex images, and retaining relevant spatial cues as outputs that can be compared with the original image.

Computational approaches have progressed from ultrasound texture analysis to radiomics and deep convolutional models, providing a technical foundation for reducing variation in manual interpretation \citep{yeh2003fibrosis,guo2024radiomics,lee2020ultrasound}. SFibAI specifically represents ultrasound grading of \emph{S. japonicum}-associated liver fibrosis using 36 ordered score levels from 0.0 to 3.5 at intervals of 0.1, while retaining their correspondence to the four clinical grades \citep{xu2026sfibai}. This representation provides a more detailed description of severity than coarse categories and establishes a basis for learning fine-grained scores directly from images. We retain this task representation and grading loss, use SFibAI as the direct baseline, and investigate how anatomical views and local regions can be organized and used within the scoring process.

An accurate score derived from whole-image features does not directly identify the acquisition view recognized by the model, the regions exhibiting relevant appearances, or how those regions contribute to the final prediction. Global representations can contain this information, but a score alone leaves users unable to inspect view predictions and regional responses separately. Acquisition-view labels and local lesion annotations provide complementary supervision: the former describes anatomical context, whereas the latter identifies locations of interest. Using these signals for grading requires an explicit computational connection between anatomical context, local features, and image-level judgment, while accommodating annotations that cover only part of an abnormal region.

Multitask learning can jointly learn grading and spatial tasks through shared representations \citep{caruana1997multitask}, but representation sharing alone does not specify how predicted views and localization outputs enter the scoring computation. Post-hoc attribution methods such as Grad-CAM can highlight regions associated with a prediction \citep{selvaraju2017gradcam}, while generating these visualizations leaves the original scoring pathway unchanged. We therefore ask: \emph{can acquisition views and local positions be organized as regional evidence that directly participates in prediction, improving fine-grained grading while retaining spatial outputs for human inspection?} Our design principle is to introduce view conditioning before spatial aggregation, allowing local features to form grading evidence within their anatomical context before combining them with whole-image judgment.

\begin{figure}[!t]
\centering
\includegraphics[width=\linewidth]{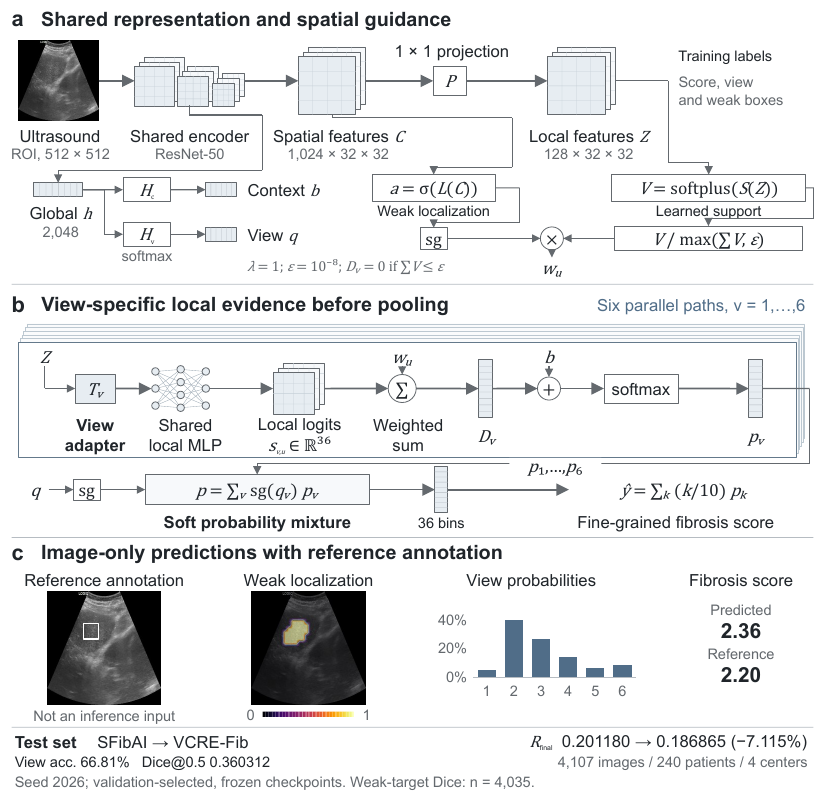}
\caption{\lang{\textbf{VCRE-Fib: view conditioning before spatial aggregation.} (a) A shared encoder supplies global context, view probabilities, weak localization and local features. (b) Six view-conditioned paths produce regional logit corrections, weighted by localization and learned support. Conditional distributions share context logits and are mixed by detached view probabilities. (c) One test case and test results from frozen, validation-selected checkpoints. Feature cells are schematic; $\sg$ denotes stop-gradient. Reference boxes are not inference inputs, and localization does not represent exhaustive segmentation.}{\textbf{VCRE-Fib：在空间聚合之前引入切面条件。}(a) 共享编码器提供全局上下文、切面概率、弱定位及局部特征。(b) 六条切面条件路径形成由定位与学习到的支持权重加权的区域 logit 修正。各条件分布共享上下文 logits，并依据截断梯度的切面概率融合。(c) 一个测试病例及使用经验证集选择并冻结的检查点得到的测试结果。特征格仅作示意，$\sg$ 表示停止梯度。参考框不参与推理，定位不表示完整病灶分割。}}\label{fig:architecture}
\end{figure}

To implement this design, we propose VCRE-Fib (View-Conditioned Regional Evidence for Fibrosis Grading), illustrated in Figure~\ref{fig:architecture}. The framework preserves a global scoring pathway and uses view labels and local lesion boxes to learn view prediction and weak localization. It generates view-conditioned regional grading evidence before spatial pooling, uses weak localization to guide regional aggregation, adds the resulting correction to the global prediction, and mixes the conditional outputs according to predicted view probabilities. This structure explicitly connects the learning, use, and presentation of spatial information: view predictions determine the mixture of conditional outputs, localization responses participate in regional aggregation, and both remain available as inspectable outputs. Inference requires only an image and returns a fibrosis score, an acquisition view, and a localization map.

We compare VCRE-Fib with SFibAI and two independently trained module-deletion variants under a common data split and evaluation protocol. On a test set of 4,107 images from 240 patients across four centers, the full model achieves the lowest prespecified composite risk in this comparison ($\risk=0.186865$), a 7.115\% reduction relative to SFibAI, together with the lowest image-level, patient-max, patient-median, and center-balanced COR. The module-deletion comparison also shows that lower grading risk does not coincide with uniformly better spatial-task metrics, motivating joint assessment of scoring performance and spatial output quality. These results support further investigation of view-conditioned regional evidence in this task and provide a concrete framework linking fine-grained grading with inspectable spatial information. Its practical benefit for clinician review and clinical decisions remains to be evaluated.

\section{Related Work}\label{sec:related}
\paragraph{Schistosomiasis-related fibrosis and ultrasound grading.}
Schistosome infection and established organ morbidity require complementary assessment: parasitological, serological, and molecular tests address infection, while imaging characterizes hepatosplenic abnormalities \citep{weerakoon2015diagnosis,sah2015imaging}. Liver biopsy provides histological information but is invasive, motivating non-invasive assessment; the limitations of fibrosis tests also require care when interpreting their outputs \citep{bravo2001biopsy,patel2020limitations}. The target in this study is an expert-assigned ultrasound appearance score.

For \emph{S. japonicum}, the anatomical distribution of fibrosis is particularly relevant. A longitudinal cohort linked ultrasound-detectable parenchymal fibrosis to infection history and documented limited reversal of severe changes after treatment \citep{carlton2010morbidity}. The Basel protocol distinguishes portal and interseptal patterns when assessing Asian schistosomiasis \citep{richter2025basel}. Together with the imaging literature \citep{sah2015imaging}, these findings motivate preserving anatomical context and visible regional appearances when predicting severity.

Computational assessment has progressed from B-mode texture features and support vector machines \citep{yeh2003fibrosis} to multiparametric ultrasomics \citep{li2019ultrasomics}, schistosomiasis-focused radiomics \citep{guo2024radiomics}, and convolutional classification with METAVIR-derived targets \citep{lee2020ultrasound}. SFibAI supplies the fine-grained score representation used here \citep{xu2026sfibai}. Acquisition, labels, and cohorts differ across these studies. Our direct comparison therefore uses SFibAI on the same cohort splits, while investigating how view and lesion-location cues can participate in grading.

\paragraph{Anatomical context and regional evidence learning.}
Multitask learning shares representations across related objectives \citep{caruana1997multitask}; SonoNet demonstrates standard-plane recognition and anatomical localization in fetal ultrasound \citep{baumgartner2017sononet}. Regional aggregation offers a complementary way to connect local information to an image-level decision. Attention-based MIL learns aggregation weights \citep{ilse2018attention}; pathology methods extend this idea with instance clustering in CLAM \citep{lu2021clam}, instance relationships in TransMIL \citep{shao2021transmil}, and additive local contributions in Additive MIL \citep{javed2022additive}. FiLM provides a general mechanism for conditioning feature transformations \citep{perez2018film}. VCRE-Fib draws on shared learning, conditioning, and aggregation to interpret regional features in their acquisition-view context before combining them with global judgment. Its regions belong to one ultrasound image, and its supervised view and localization outputs also enter the grading computation.

\paragraph{Weak localization and inspectable predictions.}
BoxSup and BoxInst learn segmentation from boxes without pixel-level masks \citep{dai2015boxsup,tian2021boxinst}. Here, boxes mark local abnormal patches and may not enclose the entire lesion, so responses beyond them remain candidates for clinical review. Concept bottleneck models expose supervised intermediate concepts \citep{koh2020concept}; VCRE-Fib exposes view and localization outputs while retaining a global route. Grad-CAM, Grad-CAM++, and integrated gradients provide post-hoc attribution approaches \citep{selvaraju2017gradcam,chattopadhay2018gradcampp,sundararajan2017axiomatic}. Our trained localization participates in regional aggregation, but this role alone does not establish faithful attribution. Saliency sanity checks motivate evaluating explanations beyond visual plausibility \citep{adebayo2018sanity}. The present clinician-annotation comparisons support inspection of spatial correspondence.

\section{VCRE-Fib: View-Conditioned Regional Evidence}\label{sec:method}
\subsection{Task and design overview}
Given an ultrasound ROI image $x$, VCRE-Fib predicts a severity distribution, acquisition view, and weak-localization map. Training uses grading labels $y$, six-class view labels, and lesion-location annotations; inference requires neither view nor box annotations. We retain the SFibAI score grid $g_k=k/10$, $k=0,\ldots,35$, and four-grade thresholds of 0.5, 1.5, and 2.5 \citep{xu2026sfibai}. The design in Figure~\ref{fig:architecture} learns spatial cues, forms regional evidence in its view context, and combines that evidence with global judgment.

\subsection{Global judgment and spatial cues}
A global path preserves whole-image context so that regional evidence complements an overall prediction. A ResNet-50 encoder \citep{he2016resnet} produces $h\in\mathbb{R}^{2048}$ and layer-3 features $C\in\mathbb{R}^{1024\times H\times W}$. With $512\times512$ inputs, $H=W=32$. The context logits are $b=H_c(h)\in\mathbb{R}^{36}$, view probabilities are $q=\softmax(H_v(h))\in\mathbb{R}^{6}$, and weak localization is $a=\sigma(L(C))$. The view and localization predictions are learned through auxiliary supervision and retained as spatial outputs for inspection.

Local boxes identify clinician-marked regions without specifying exhaustive contours. The implemented weak-box objective combines an inside-response term with a constraint on the outside response fraction. It guides spatial learning but supplies no complete lesion-contour target. Whether learned responses extend beyond a local annotation and correspond to broader pathology is therefore an empirical question examined through actual cases.

\subsection{Conditioning before regional aggregation}
Local features are interpreted in their anatomical context. A $1\times1$ projection produces $Z=P(C)\in\mathbb{R}^{128\times H\times W}$. Six view adapters $T_v$, each a $128\rightarrow128$ pointwise convolution and GELU, share a $128\rightarrow64\rightarrow36$ MLP $f$:
\begin{equation}
s_{v,u}=f(T_v(Z)_u),\quad v=1,\ldots,6,\quad u=1,\ldots,HW .
\label{eq:local}
\end{equation}
Conditioning occurs before pooling and thus changes the grading evidence at each location. The vector $s_{v,u}$ is a local logit contribution candidate, not a separately calibrated regional severity score.

\subsection{Localization-guided regional grading}
Weak localization supplies spatial relevance, while a support head $V=\operatorname{softplus}(S(Z))$ learns aggregation weights through grading. They organize the regional correction:
\begin{equation}
c_{v,u}=\lambda\frac{V_u\sg(a_u)}{\max(\sum_{u'}V_{u'},\epsilon)}s_{v,u},
\qquad D_v=\sum_u c_{v,u}.
\label{eq:regional}
\end{equation}
The reference configuration uses $\lambda=1$ and $\epsilon=10^{-8}$, with zero correction when total support is at most $\epsilon$. Normalization by $\sum V$ allows uniformly low localization responses to attenuate the correction. The operator $\sg$ stops gradients at the indicated interface.

Regional evidence is added to global context, and view probabilities retain uncertainty across conditional judgments:
\begin{equation}
p_v=\softmax(b+D_v),\qquad
p=\sum_v\sg(q_v)p_v,\qquad
\hat y=\sum_k g_kp_k.
\label{eq:mixture}
\end{equation}
The model mixes conditional probabilities. Localization therefore participates in the grading path while remaining an explicit output.

\subsection{Supervision and inference outputs}
Following shared-representation multitask learning \citep{caruana1997multitask}, the training objective is
\begin{equation}
\mathcal L=\mathcal L_{\mathrm{grade}}(p,y)+0.1\mathcal L_{\mathrm{view}}+0.1\mathcal L_{\mathrm{box}}.
\label{eq:loss}
\end{equation}
View prediction uses cross-entropy; the weak-box loss uses valid local boxes, as specified in Appendix~\ref{sec:appendix-implementation}. The grading interface consumes $\log p$. Gradients are stopped at the grading uses of $q$ and $a$, while auxiliary losses still update the encoder and grading still trains the context, local-evidence, and support paths. VCRE-Fib uses the same hybrid grading-loss formulation and weights as SFibAI: both compute MSE and the clinical-grade boundary penalty using the real-valued scores $\hat y=0.1\sum_k kp_k$ and $y=0.1j$, where $j$ is the target bin. Appendix~\ref{sec:appendix-implementation} gives the common grading-loss formulation.

Inference returns $\hat y$, the view $\arg\max_v q_v$, and localization $a$. Side-by-side original images, clinician annotations, and predictions expose the spatial outputs available for review. Output completeness here means retaining both view and regional information alongside the score; it does not mean complete lesion coverage.

\section{Experimental Design}\label{sec:setup}
\paragraph{Data and supervision.}
Our cohort was re-curated from the institutional ultrasound database described by \citet{xu2026sfibai}. The prediction target is the existing expert-assigned ultrasound appearance score. We cleaned and organized these scores and the existing clinician-drawn lesion boxes, and recovered six-class view labels from acquisition-plane identifiers. The cohort contains 108,709 ROI images from 6,373 patients across 35 centers. Training, validation, and test comprise 83,722/4,906, 20,880/1,227, and 4,107/240 images/patients, respectively. Patients do not overlap across splits; all four test centers are represented in training. All models use the same grading data and splits, with spatial supervision determined by retained modules. Boxes at the highest recorded score form weak targets, not exhaustive contours. All 4,107 test images have view annotations, and 4,035 meet weak-target evaluation criteria. Appendix~\ref{sec:appendix-cohort} describes annotation provenance; Table~\ref{tab:views} lists view names and counts.

\paragraph{Compared models.}
We compare SFibAI, full VCRE-Fib, and two deletion variants (Table~\ref{tab:matrix}). \emph{w/o View} retains weak localization and regional grading but removes the view head, view-specific adapters, posterior mixture, and view loss. \emph{w/o Weak Loc.} retains view-conditioned grading but removes the localization head and weak-box loss, setting $a_u=1$ in Equation~\ref{eq:regional}. Both retain the support and global paths and are trained independently, rather than fine-tuned from the full model. The contrasts evaluate whole model designs; they do not isolate supervision, capacity, and information routing from one another.
\begin{table}[!htbp]
\centering\normalsize
\caption{\lang{Compared models and inference outputs. All use image-only inference, the same backbone initialization and grading objective. Deletion variants retain the regional grading path and are trained independently.}{比较模型及推理输出。均只输入图像，骨干初始化与评分目标一致；删除变体保留区域评分路径，并独立训练。}}\label{tab:matrix}
\setlength{\tabcolsep}{3pt}
\fitresulttable{%
\begin{tabular}{@{}llll@{}}
\toprule
Method & View module & Weak-localization module & Outputs \\
\midrule
SFibAI & Absent & Absent & Score \\
w/o View & Removed & Retained & Score + region \\
w/o Weak Loc. & Retained & Removed & Score + view \\
VCRE-Fib & Retained & Retained & Score + view + region \\
\bottomrule
\end{tabular}}
\end{table}

\paragraph{Training, selection, and freezing.}
All four use the same random seed (2026), ResNet-50, ImageNet-1K-V2 initialization, $512\times512$ inputs, global batch 24, bfloat16 training, AdamW \citep{loshchilov2019adamw}, and the same real-score hybrid grading objective. All methods, including full VCRE-Fib, use a common 90-epoch cap for the training comparison. The reported trajectories cover epochs 1--90, and validation selects from epochs 21--90 by lower $\risk$, lower image COR, then earlier epoch. The common recipe is specified in Appendix~\ref{sec:appendix-implementation}. Checkpoints, configurations, and the endpoint are frozen before test comparison; test determines the observed ranking.

\paragraph{Evaluation questions.}
We ask whether each spatially informed variant improves on the baseline, how jointly using the modules changes grading and spatial outputs, and what information the outputs expose. The lower-is-better primary endpoint is
\begin{equation}
\risk=0.4R_{\mathrm{image}}+0.4R_{\mathrm{patient\text{-}max}}+0.2R_{\mathrm{center}} .
\label{eq:risk}
\end{equation}
Each component uses the locked clinical objective risk (COR); the center term weights per-center patient-max COR by square-root patient counts. Patient-median is reported separately. Differences describe observed results, without significance testing. Bold table values indicate the best unrounded result per metric, including exact ties.

\paragraph{Spatial outputs and visualization.}
View metrics include accuracy/macro-F1; weak-target Dice/IoU use threshold 0.5. Four selected test cases spanning four true views illustrate the outputs, with two validation examples in the appendix. Case-selection criteria, patient deduplication, and common display settings are specified in Appendix~\ref{sec:appendix-case-protocol}.

\section{Grading and Spatial Prediction Results}\label{sec:results}
\subsection{Test set primary results}\label{sec:test}
We tested whether spatial information improves fine-grained grading. On the 4,107-image, 240-patient, four-center test set, full VCRE-Fib achieves the lowest prespecified composite risk ($\risk=0.186865$), followed by w/o Weak Loc.\ (0.191308), w/o View (0.193799), and SFibAI (0.201180; Table~\ref{tab:main}). Their respective absolute differences from SFibAI are $-0.014315$, $-0.009872$, and $-0.007380$, corresponding to reductions of 7.115\%, 4.907\%, and 3.668\%, computed before rounding. All spatial variants outperform SFibAI on this endpoint. Results use validation-selected, frozen checkpoints.
\begin{table}[!htbp]
\centering\normalsize
\caption{\lang{Test set primary results: 4,107 images, 240 patients, four centers. All checkpoints were frozen after validation selection. All risks are lower-is-better. Bold denotes the best unrounded value per metric; exact ties are all bold.}{Test set 主结果：4,107 图、240 患者、4 中心；全部检查点经验证集选择后冻结，风险均越低越好。各指标按未取整数值加粗最优项，精确并列者均加粗。}}\label{tab:main}
\setlength{\tabcolsep}{3pt}
\fitresulttable{%
\begin{tabular}{@{}rlrrrrr@{}}
\toprule
Rank & Method & $R_{\mathrm{final}}$ & Image COR & Max COR & Median COR & Center COR \\
\midrule
1 & VCRE-Fib & \textbf{0.186865} & \textbf{0.177908} & \textbf{0.200528} & \textbf{0.154344} & \textbf{0.177453} \\
2 & w/o Weak Loc. & 0.191308 & 0.186343 & 0.202422 & 0.165844 & 0.179008 \\
3 & w/o View & 0.193799 & 0.181518 & 0.210002 & 0.158449 & 0.185957 \\
4 & SFibAI & 0.201180 & 0.185938 & 0.219742 & 0.162587 & 0.194539 \\
\bottomrule
\end{tabular}}
\end{table}

\subsection{Module deletions reveal distinct risk profiles}
Independently trained deletion variants show distinct advantages. The view-retaining w/o Weak Loc.\ has lower patient-max COR, whereas the localization-retaining w/o View has lower image and patient-median COR (Table~\ref{tab:main}). The full model leads all three risks and center-balanced COR, improving composite risk by 0.004443 over the better variant, w/o Weak Loc. Gains thus span risk measures. These design-level comparisons do not isolate supervision, capacity, or information use, or establish statistical superiority or module synergy.

\subsection{Grading gains extend across images and patients}
Specific grading metrics improve alongside composite risk. Relative to SFibAI, full VCRE-Fib reduces image MAE from 0.391313 to 0.377857, increases four-grade accuracy from 64.11\% to 67.03\%, and reduces severe errors from 9.98\% to 9.37\% (Table~\ref{tab:image}). The full model also leads the remaining image metrics: RMSE, TMAE, COR, $\pm0.3$ and $\pm0.5$ accuracy, and macro-F1. The observed gains therefore span score error, agreement within tolerances, and four-grade discrimination.
\begin{table}[!htbp]
\centering\normalsize
\caption{\lang{Image-level test metrics; 4,107 images per model. Bold denotes the best unrounded value per metric; exact ties are all bold.}{图像级测试集指标，每模型 4,107 图。各指标按未取整数值加粗最优项，精确并列者均加粗。}}\label{tab:image}
\setlength{\tabcolsep}{3pt}
\fitresulttable{%
\begin{tabular}{@{}lrrrr@{}}
\toprule
Metric & SFibAI & w/o View & w/o Weak Loc. & VCRE-Fib \\
\midrule
MAE $\downarrow$ & 0.391313 & 0.384734 & 0.392205 & \textbf{0.377857} \\
RMSE $\downarrow$ & 0.568137 & 0.564076 & 0.567780 & \textbf{0.553435} \\
Acc.$\pm0.3$ (\%) $\uparrow$ & 56.17 & 57.51 & 56.05 & \textbf{57.95} \\
Acc.$\pm0.5$ (\%) $\uparrow$ & 69.61 & 70.08 & 69.47 & \textbf{70.15} \\
4-grade acc. (\%) $\uparrow$ & 64.11 & 66.13 & 64.13 & \textbf{67.03} \\
Macro-F1 $\uparrow$ & 0.639702 & 0.663674 & 0.643722 & \textbf{0.671466} \\
TMAE $\downarrow$ & 0.129177 & 0.127592 & 0.129382 & \textbf{0.123100} \\
Severe error (\%) $\downarrow$ & 9.98 & 10.03 & 9.76 & \textbf{9.37} \\
COR $\downarrow$ & 0.185938 & 0.181518 & 0.186343 & \textbf{0.177908} \\
\bottomrule
\end{tabular}}
\end{table}

The full model also has the lowest patient-max, patient-median, and center-balanced COR (Table~\ref{tab:main}). Some thresholded accuracies favor deletion variants; Appendix~\ref{sec:appendix-patient} reports the complete patient-level metrics and these differences.

\subsection{Lower grading risk accompanies spatial-task trade-offs}
We next examined the accompanying spatial predictions. Full VCRE-Fib reaches view accuracy of 66.81\%, macro-F1 of 0.665861, and CE of 0.916542 on 4,107 test images, with weak-target Dice/IoU of 0.360312/0.235487 on 4,035 eligible images (Table~\ref{tab:auxiliary}). Yet w/o Weak Loc.\ achieves higher view accuracy and macro-F1 (75.70\% and 0.757500), and w/o View achieves higher Dice/IoU (0.430661/0.298825). Thus, the lowest grading risk does not coincide with the best spatial-task scores. Only the full model provides all three outputs, with auxiliary-task trade-offs. Weak-target overlap measures agreement with coarse local annotations, not complete lesion segmentation.
\begin{table}[!htbp]
\centering\normalsize
\caption{\lang{Spatial test outputs. N/A indicates an absent head. Weak-target N counts eligible images, not rectangles; Dice/IoU measure agreement with local annotations, not complete lesion segmentation. Bold denotes the best unrounded value per metric; exact ties are all bold.}{空间输出测试集结果，缺少对应头记 N/A。弱定位 N 是有效图像数而非矩形框数；Dice/IoU 衡量与局部弱标注的一致性，不代表完整病灶分割。各指标按未取整数值加粗最优项，精确并列者均加粗。}}\label{tab:auxiliary}
\setlength{\tabcolsep}{3pt}
\fitresulttable{%
\begin{tabular}{@{}lrrrr@{}}
\toprule
Metric & SFibAI & w/o View & w/o Weak Loc. & VCRE-Fib \\
\midrule
View N & \na & \na & 4107 & 4107 \\
View acc. (\%) $\uparrow$ & \na & \na & \textbf{75.70} & 66.81 \\
View macro-F1 $\uparrow$ & \na & \na & \textbf{0.757500} & 0.665861 \\
View CE $\downarrow$ & \na & \na & \textbf{0.690513} & 0.916542 \\
Weak-target N & \na & 4035 & \na & 4035 \\
Dice@0.5 $\uparrow$ & \na & \textbf{0.430661} & \na & 0.360312 \\
IoU@0.5 $\uparrow$ & \na & \textbf{0.298825} & \na & 0.235487 \\
\bottomrule
\end{tabular}}
\end{table}

\subsection{Joint outputs expose grading--spatial disagreements}
\begin{figure}[!tb]
\centering
\includegraphics[width=\linewidth]{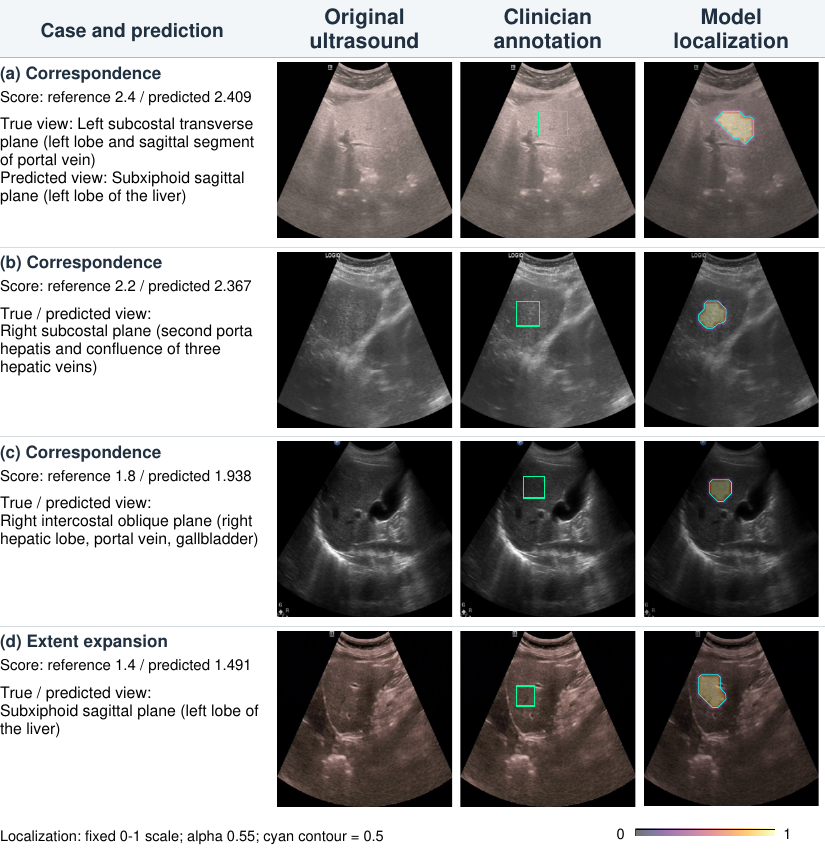}
\caption{\lang{Four selected test cases, with reference and predicted scores and acquisition views at left. Predictions use the frozen checkpoint evaluated quantitatively. Rows (a)--(c) show localization correspondence; (d) shows a response beyond local boxes. Green boxes denote clinician annotations; localization uses a $[0,1]$ scale, opacity 0.55, and a cyan 0.5 contour. In row (a), the predicted view differs from the reference. These selected examples illustrate spatial outputs without establishing population localization performance or complete lesion extent.}{四个筛选的测试集病例，左侧标出参考及预测评分和扫查切面。预测使用参与定量评价的冻结检查点。(a)--(c) 展示定位对应，(d) 展示局部框外响应。绿色框为医生标注；定位色标为 $[0,1]$、透明度为 0.55，青色轮廓阈值为 0.5。(a) 的预测切面与参考切面不同。这些筛选示例用于展示空间输出，不代表总体定位水平或完整病灶范围。}}
\label{fig:cases}
\end{figure}

Figure~\ref{fig:cases} illustrates four test cases spanning distinct true views, juxtaposing ultrasound, clinician annotation, and localization with reference/predicted scores and full view names. Rows (a)--(c) show localization correspondence. In (a), scores agree closely (2.4/2.409), yet the true left subcostal transverse view is classified as subxiphoid sagittal. Close score agreement thus need not imply a correct spatial judgment; the joint outputs make this disagreement visible.

Row (d), with scores 1.4/1.491, shows responses outside local rectangles under the fixed scale and threshold; two validation examples appear separately in the appendix. Responses outside annotations offer locations to inspect but may represent relevant tissue or erroneous activation, establishing neither complete lesion coverage nor clinical correctness. All qualitative case predictions were generated with the frozen checkpoint evaluated quantitatively. They illustrate inspectable information, without assigned clinical interpretations or claims of evaluated clinical-review benefit.

\subsection{Computational cost of joint prediction}
We quantified the cost of providing grading and both spatial outputs. Matched FP32 measurements on an A100-SXM4-80GB yield median batch-one latencies of 5.620, 6.171, 7.122, and 7.371 ms for SFibAI, w/o View, w/o Weak Loc., and full VCRE-Fib. The full model adds 7.81\% parameters, 4.24\% counted MACs, and 1.751 ms median latency over SFibAI. Appendix Table~\ref{tab:resources} reports latency, P95, and memory at batch sizes one and eight.

\section{Discussion}\label{sec:discussion}
Grading risk, view accuracy, and weak-target overlap assess severity prediction, acquisition-view recognition, and agreement with local annotations, respectively. The full model's lowest grading risk and the deletion variants' higher spatial-task scores motivate evaluating the score and each accompanying output together. The module-deletion comparisons evaluate whole designs rather than isolating supervision, capacity, or information use, leaving the mechanisms behind these trade-offs unresolved.

Local rectangular boxes provide positional cues for spatial learning. Comparing regional maps with the original image and clinician boxes makes correspondence to these cues visible and identifies additional regions for inspection. Weak-target overlap measures positional agreement, whereas complete lesion coverage is a separate objective. Responses outside local boxes may reflect relevant tissue or erroneous activation; more complete spatial references and clinician review could distinguish these possibilities.

For clinical review, the joint outputs provide three objects to inspect: the predicted view, regional response, and final score. The illustrated view error despite close score agreement shows how these outputs expose different aspects of a prediction. Displaying them alongside the original image would allow separate checks of anatomical context and local responses. Their practical benefit remains to be evaluated through reader studies of error identification, appropriate assessment revisions, and review time.

Deployment studies should measure preprocessing, inference, and display together on target hardware, beyond the model-only costs reported here. Independent-cohort evaluation should test grading and spatial outputs across different patient populations and acquisition settings.

\section{Conclusion}\label{sec:conclusion}
VCRE-Fib combines view-conditioned regional evidence with a global prediction for fine-grained ultrasound grading of \emph{Schistosoma japonicum}-associated liver fibrosis. In the frozen-checkpoint comparison, the full model reduced composite test risk by 7.115\% relative to SFibAI while returning score, view, and regional outputs. The full model's grading gains coexisted with higher view accuracy or weak-target overlap in the corresponding deletion variants. This trade-off makes the quality of each spatial output a distinct part of evaluating the joint system, alongside its final grading performance. The joint outputs make anatomical and regional judgments available for inspection alongside the score. Further evaluation should examine their usefulness for clinician review and their performance in independent cohorts.

\label{main-text-end}
\clearpage
\section*{Ethics Statement}\label{post-main-start}
The retrospective use of ultrasound images and associated annotations in this study was covered by institutional ethics approval. Written informed consent was obtained from all participants before enrolment.
\section*{Reproducibility Statement}
The architecture and evaluation protocol are specified in Sections~\ref{sec:method}--\ref{sec:setup}. Appendix~\ref{sec:appendix-implementation} details data preparation and the shared training setup; Appendix~\ref{sec:appendix-results} reports complete test and resource metrics, Appendix~\ref{sec:appendix-analyses} defines case selection and display, and Appendix~\ref{sec:appendix-selection} provides checkpoint selection and all four training trajectories.
\section*{AI Use Statement}
OpenAI Codex assisted with implementation inspection, manuscript drafting, translation, and editing, reference checking, document compilation, and presentation of existing results and case panels. It did not generate experimental observations. The authors are responsible for the scientific content and interpretation.
\bibliographystyle{iclr2027_conference}
\bibliography{references}
\section*{Appendix}
\addcontentsline{toc}{section}{Appendix}
\label{appendix-start}
\appendix
\setcounter{table}{0}
\setcounter{figure}{0}
\setcounter{equation}{0}
\renewcommand{\thetable}{S\arabic{table}}
\renewcommand{\thefigure}{S\arabic{figure}}
\renewcommand{\theequation}{S\arabic{equation}}
\renewcommand{\theHtable}{appendix.\arabic{table}}
\renewcommand{\theHfigure}{appendix.\arabic{figure}}
\renewcommand{\theHequation}{appendix.\arabic{equation}}

\section{Cohort, annotations, and implementation details}\label{sec:appendix-implementation}
\subsection{Cohort provenance and annotation scope}\label{sec:appendix-cohort}
\paragraph{Reference scores and cohort curation.}
The cohort was re-curated from the institutional ultrasound database described by \citet{xu2026sfibai}. We retained its expert-assigned ultrasound appearance scores, ranging from 0.0 to 3.5 in increments of 0.1. The published annotation protocol refines four clinical grades into this score scale and describes standardized instruction, centralized review, and expert review of grading disagreements. The present study cleaned and organized the existing images and annotations without regrading them. Its cohort and splits differ from those of the earlier study; the SFibAI baseline was trained and evaluated on the present splits.

\paragraph{Acquisition views and lesion-location annotations.}
Six-class view labels were recovered from the plane identifiers recorded during image acquisition and organized during data cleaning. Lesion-location supervision likewise uses the existing clinician-drawn rectangular boxes described by \citet{xu2026sfibai}, after cleaning and organization. These boxes mark image regions requiring attention and carry the original grading annotations. View labels therefore describe the recorded acquisition plane, whereas boxes identify local regions within the image. Neither annotation type is required at inference.

\paragraph{Coverage and weak-target scope.}
The curated records contain view and lesion-location annotations for every image, although not every record supplies an eligible nonempty weak target. The box objective uses valid nonempty boxes with a positive target score, as specified in Appendix~\ref{sec:appendix-implementation}. Weak-localization evaluation includes 4,035 eligible test images; view evaluation includes all 4,107 test images. Local boxes do not delineate complete lesion contours, so responses beyond them require separate assessment of spatial relevance.

\begin{table}[!htbp]
\centering
\caption{\lang{Composition of the curated cohort used in this study. Centers overlap between splits; patients do not.}{本文整理后队列的组成。中心可跨划分出现，患者不跨划分。}}\label{tab:cohort}
\normalsize
\begin{tabular}{@{}lrrrr@{}}
\toprule
 & \lang{Total}{总计} & \lang{Train}{训练} & \lang{Validation}{验证} & \lang{Test}{测试} \\
\midrule
\lang{Images}{图像} & 108,709 & 83,722 & 20,880 & 4,107 \\
\lang{Patients}{患者} & 6,373 & 4,906 & 1,227 & 240 \\
\lang{Centers}{中心} & 35 & 35 & 33 & 4 \\
\bottomrule
\end{tabular}
\end{table}

\subsection{SFibAI baseline and training implementation}
\paragraph{Identifying the SFibAI baseline.}
The method source is \citet{xu2026sfibai}. The baseline follows the authors' public implementation. The baseline uses ResNet-50, global average pooling, a linear $2048\rightarrow36$ head, softmax, and an expected score on 0.0--3.5. View and localization prediction heads are absent. Upstream permits an explicit initialization checkpoint and otherwise constructs a randomly initialized backbone; our common ImageNet initialization is an adaptation for this comparison, not an upstream default.

\paragraph{Grading objectives.}
VCRE-Fib and SFibAI use the same fibrosis-score scale and hybrid-loss weights $(\alpha,\beta,\gamma)=(1,0.02,0.02)$. For target bin $j\in\{0,\ldots,35\}$ and predicted probabilities $p_k$, both convert bin indices to real-valued clinical scores before computing MSE and the clinical-grade boundary penalty:
\[
\begin{aligned}
y&=0.1j,\qquad \hat y=0.1\sum_{k=0}^{35}kp_k,\\
L_{\mathrm{grade}}&=L_{\mathrm{KL,local}}+0.02(\hat y-y)^2
 +0.02|\hat y-y|\mathbf{1}(\hat c\ne c).
\end{aligned}
\]
Here $c$ and $\hat c$ are obtained from $y$ and $\hat y$ using thresholds 0.5, 1.5, and 2.5, with a score exactly at a threshold assigned to the higher grade. The displayed per-image terms are averaged over the batch. Both MSE and the absolute-error boundary term are evaluated on the 0.0--3.5 score scale.

The local-KL term uses a normalized Gaussian soft target over 36 bins with standard deviation one bin and a local window centered on the target bin. Both methods use this hybrid grading-loss formulation with aligned score units and weights.

\paragraph{Training protocol.}
All four models use the same random seed (2026), ImageNet-1K-V2 weights, batch size 24, bfloat16 precision, AdamW learning rate $10^{-4}$ and weight decay $10^{-4}$, and StepLR factor 0.6 every 15 epochs. They share ROI preprocessing, stretch resize, right-angle rotation/saturation augmentation, patient-disjoint splits, and the grading objective above. This standardizes the public SFibAI release's different crop/augmentation and initialization defaults. Training trajectories and checkpoint comparison use the same 90-epoch cap for SFibAI, both deletion variants, and full VCRE-Fib. Validation selects from epochs 21--90 by $\risk$, then image COR and earlier epoch, replacing the upstream tolerance-accuracy selector. The selected epochs are 65, 80, 39, and 47, respectively. This comparison on the curated cohort is not a reproduction of the original SFibAI cohort experiment.

\paragraph{Additional VCRE-Fib supervision.}
Six-class view cross-entropy has weight 0.1. The weak-box loss combines inside log-sum-exp pooling (temperature 10) and the outside response fraction with unit inner weights and overall weight 0.1. Only valid nonempty boxes with target bin $j>0$ participate; exact score 0.0 is excluded, not the entire F0 grade. Weak targets are transformed in the ROI frame. The support map is learned from grading. Auxiliary losses update the shared encoder, while $q$ and $a$ are detached only at their grading interfaces. The w/o View variant removes the view head, six adapters, posterior mixture, and view CE while retaining unconditioned regional evidence and weak localization. The w/o Weak Loc.\ variant removes the localization head and box loss, using neutral $a_u=1$ with the view and support paths retained. Each deletion variant starts independently from the prescribed initialization.

\FloatBarrier
\section{Complete test and resource reporting}\label{sec:appendix-results}
\subsection{Test set primary results}
The frozen, validation-selected checkpoints are evaluated on 4,107 images, 240 patients, and four centers. The order is VCRE-Fib, w/o Weak Loc., w/o View, and SFibAI (Table~\ref{tab:comparison}), with lower $\risk$ indicating better performance. Main Tables~\mainref{tab:main}--\mainref{tab:auxiliary} report the primary, image, and spatial outcomes; Appendix~\ref{sec:appendix-patient} gives the detailed patient-level metrics.
\begin{table}[!htbp]
\centering\normalsize
\caption{\lang{Observed test differences relative to SFibAI. Differences are method minus baseline; relative differences divide by the baseline, using unrounded values. Bold denotes the best unrounded value per metric; exact ties are all bold.}{四模型相对 SFibAI 的测试集观测差异。差值为方法减基线，相对差除以基线，均用未取整数值计算。各指标按未取整数值加粗最优项，精确并列者均加粗。}}\label{tab:comparison}
\setlength{\tabcolsep}{3pt}
\fitresulttable{%
\begin{tabular}{@{}lrrrr@{}}
\toprule
Method & Rank & $R_{\mathrm{final}}\downarrow$ & $\Delta$ absolute $\downarrow$ & $\Delta$ relative (\%) $\downarrow$ \\
\midrule
VCRE-Fib & 1 & \textbf{0.186865} & \textbf{-0.014315} & \textbf{-7.115} \\
w/o Weak Loc. & 2 & 0.191308 & -0.009872 & -4.907 \\
w/o View & 3 & 0.193799 & -0.007380 & -3.668 \\
SFibAI & 4 & 0.201180 & 0.000000 & 0.000 \\
\bottomrule
\end{tabular}}
\end{table}

\paragraph{Metric definitions.}
Let $e=|\hat y-y|$ and let $d$ be the absolute difference in four-level grades with thresholds 0.5, 1.5, and 2.5. The locked per-item risk is
\[
\operatorname{COR}=0.35e/3.5+0.15\mathbf{1}(e>0.3)+0.15\mathbf{1}(e>0.5)+0.25d/3+0.10\mathbf{1}(d\ge2).
\]
Patient-max and patient-median aggregate reference and predicted scores separately. Center-balanced COR uses square-root patient-count weights. TMAE is $\operatorname{mean}\max(e-0.5,0)$ and severe-error rate is $\operatorname{mean}\mathbf{1}(e>1.0)$, distinct from the two-grade error term inside COR.
\subsection{Detailed patient-level grading}\label{sec:appendix-patient}
Under patient-max and patient-median aggregation, the full model retains the lowest COR (0.200528 and 0.154344), alongside the lowest median MAE (0.339273; Table~\ref{tab:patient}) and center-balanced COR (0.177453; main-text Table~\mainref{tab:main}). Some thresholded accuracies still favor variants: w/o Weak Loc.\ has higher patient-max $\pm0.5$ accuracy (67.50\%), and w/o View has higher median four-grade accuracy (72.08\%). The full model's advantage thus spans overall risk and multiple grading measures.
\begin{table}[!htbp]
\centering\normalsize
\caption{\lang{Patient-level test metrics, 240 patients per row. Reference and prediction are aggregated separately; accuracies and severe errors are percentages. Bold compares models within each aggregation.}{患者级测试集指标，每行 240 患者；参考与预测分别聚合，准确率和严重错误率以百分数表示。加粗在同一聚合层级内比较。}}\label{tab:patient}
\setlength{\tabcolsep}{2pt}
\fitresulttable{%
\begin{tabular}{@{}llrrrrrrr@{}}
\toprule
Method & Level & MAE $\downarrow$ & \shortstack{Acc.\\$\pm.3\uparrow$} & \shortstack{Acc.\\$\pm.5\uparrow$} & 4-grade $\uparrow$ & TMAE $\downarrow$ & Severe $\downarrow$ & COR $\downarrow$ \\
\midrule
SFibAI & Max & 0.472421 & 50.42 & 63.75 & 61.25 & 0.182625 & 16.25 & 0.219742 \\
w/o View & Max & 0.445159 & \textbf{53.75} & 66.25 & 63.75 & 0.171660 & 15.42 & 0.210002 \\
w/o Weak Loc. & Max & 0.427688 & 53.33 & \textbf{67.50} & 63.75 & 0.149615 & 12.50 & 0.202422 \\
VCRE-Fib & Max & \textbf{0.420558} & \textbf{53.75} & 65.42 & \textbf{65.42} & \textbf{0.143967} & \textbf{11.67} & \textbf{0.200528} \\
\midrule
SFibAI & Median & 0.348790 & 62.08 & 72.08 & 67.08 & 0.102808 & 7.92 & 0.162587 \\
w/o View & Median & 0.346292 & 61.25 & 73.75 & \textbf{72.08} & 0.108306 & 7.92 & 0.158449 \\
w/o Weak Loc. & Median & 0.361912 & 59.58 & 72.92 & 69.58 & 0.113333 & 9.17 & 0.165844 \\
VCRE-Fib & Median & \textbf{0.339273} & \textbf{62.92} & \textbf{74.58} & 71.67 & \textbf{0.102352} & \textbf{7.50} & \textbf{0.154344} \\
\bottomrule
\end{tabular}}
\end{table}

\subsection{Additional grading metrics and center-level results}
Per-center estimates are descriptive; one center has two patients. Test/val exports retain per-grade support/recall and all COR terms. Four-grade probabilities sum the 36 bins; image AUPRC uses average precision and ECE uses ten bins. Class-wise AUROC/AUPRC are reported as N/A when either positive or negative examples are absent. Patient probabilities are not aggregated.
\begin{table}[!htbp]
\centering\normalsize
\caption{\lang{Additional test grading and image-only probability metrics. Bold denotes the best unrounded value per metric; exact ties are all bold.}{补充测试集分级指标与仅图像级的概率指标。各指标按未取整数值加粗最优项，精确并列者均加粗。}}\label{tab:additional}
\setlength{\tabcolsep}{3pt}
\fitresulttable{%
\begin{tabular}{@{}lrrrr@{}}
\toprule
Metric & SFibAI & w/o View & w/o Weak Loc. & VCRE-Fib \\
\midrule
Patient Max RMSE $\downarrow$ & 0.665765 & 0.642874 & 0.611087 & \textbf{0.596914} \\
Patient Max macro-F1 $\uparrow$ & 0.606287 & 0.635779 & 0.636286 & \textbf{0.655596} \\
Patient Median RMSE $\downarrow$ & 0.508183 & 0.519102 & 0.528079 & \textbf{0.505147} \\
Patient Median macro-F1 $\uparrow$ & 0.675454 & \textbf{0.726897} & 0.702926 & 0.721044 \\
\midrule
Image macro-AUROC $\uparrow$ & 0.879791 & 0.886155 & 0.879378 & \textbf{0.891698} \\
Image macro-AUPRC $\uparrow$ & 0.699348 & 0.719365 & 0.699530 & \textbf{0.729264} \\
Image multiclass Brier $\downarrow$ & 0.526870 & 0.499022 & 0.515159 & \textbf{0.474036} \\
Image ECE $\downarrow$ & 0.206199 & 0.181620 & 0.177804 & \textbf{0.164196} \\
\bottomrule
\end{tabular}}
\end{table}

\begin{table}[!htbp]
\centering\normalsize
\caption{\lang{Per-center test patient-max COR (lower is better). Bold compares methods within a center; small-center estimates remain descriptive.}{逐中心测试集 patient-max COR（越低越好），加粗在同一中心内比较；小中心估计仅作描述。}}\label{tab:centers}
\setlength{\tabcolsep}{3pt}
\fitresulttable{%
\begin{tabular}{@{}lrrrrrr@{}}
\toprule
Center & Images & Patients & SFibAI & w/o View & w/o Weak Loc. & VCRE-Fib \\
\midrule
center\_05 & 25 & 2 & 0.005361 & 0.006614 & 0.005765 & \textbf{0.005329} \\
center\_07 & 1948 & 115 & 0.201563 & 0.192100 & 0.192990 & \textbf{0.183879} \\
center\_15 & 1599 & 93 & 0.308194 & 0.295332 & \textbf{0.274507} & 0.281672 \\
center\_17 & 535 & 30 & 0.029521 & 0.027661 & 0.028223 & \textbf{0.025816} \\
\bottomrule
\end{tabular}}
\end{table}

\FloatBarrier

\subsection{Resource measurements}
\label{sec:appendix-resources}
\begin{table}[!htbp]
\centering\normalsize
\caption{\lang{Matched resource measurements; latency is milliseconds per batch and memory is peak allocated MiB. All quantities are lower-is-better; bold compares models within each batch size.}{统一条件下的资源测量：延迟为毫秒/批次，显存为峰值已分配 MiB。数值均越低越好，加粗在同一 batch size 内比较。}}\label{tab:resources}
\setlength{\tabcolsep}{3pt}
\fitresulttable{%
\begin{tabular}{@{}lrrrrrr@{}}
\toprule
Method & Batch & Parameters & GMACs/image & Median ms & P95 ms & Peak MiB \\
\midrule
SFibAI & 1 & \textbf{23581796} & \textbf{21.353} & \textbf{5.620} & \textbf{5.626} & \textbf{154.318} \\
w/o View & 1 & 25311562 & 22.103 & 6.171 & 6.188 & 160.919 \\
w/o Weak Loc. & 1 & 24832975 & 21.654 & 7.122 & 7.165 & 159.093 \\
VCRE-Fib & 1 & 25422928 & 22.258 & 7.371 & 7.397 & 161.344 \\
\midrule
SFibAI & 8 & \textbf{23581796} & \textbf{21.353} & \textbf{30.088} & \textbf{30.119} & \textbf{539.318} \\
w/o View & 8 & 25311562 & 22.103 & 31.067 & 31.133 & 545.919 \\
w/o Weak Loc. & 8 & 24832975 & 21.654 & 32.338 & 32.442 & 544.093 \\
VCRE-Fib & 8 & 25422928 & 22.258 & 32.800 & 32.979 & 546.344 \\
\bottomrule
\end{tabular}}
\end{table}

All four frozen models were measured on the same NVIDIA A100-SXM4-80GB with PyTorch 2.11.0+cu128, FP32, TF32 disabled, and $512\times512$ inputs. Fifty warm-up iterations preceded 200 synchronized wall-time measurements. Latency is per batch, excludes loading/transfer, and covers the model forward with inputs already on the GPU. Peak allocated memory is reported in MiB. Counted Conv2d/Linear operations exclude elementwise, pooling, and normalization; FLOPs are twice the reported MACs.

\FloatBarrier
\section{Spatial predictions and qualitative examples}\label{sec:appendix-analyses}
\subsection{View names and weak-target summaries}
\begin{table}[!htbp]
\centering\normalsize
\caption{\lang{Six acquisition views and test recall (percent). Bold compares the two view-capable models within each class. Main-paper cases cover four true views.}{六类扫查切面与测试集召回率（百分数），在每类的两个含切面头模型中加粗最优项；正文病例覆盖四种真实切面。}}\label{tab:views}
\setlength{\tabcolsep}{3pt}
\fitresulttable{%
\begin{tabular}{@{}lp{.49\linewidth}rrr@{}}
\toprule
View & Anatomical/acquisition view & N & w/o Weak Loc. & VCRE-Fib \\
\midrule
1 & Subxiphoid sagittal plane (left lobe of the liver) & 694 & \textbf{93.66} & 84.73 \\
2 & Left subcostal transverse plane (left lobe and sagittal segment of portal vein) & 699 & \textbf{74.82} & 65.09 \\
3 & Right subcostal plane (second porta hepatis and confluence of three hepatic veins) & 689 & \textbf{59.51} & 54.57 \\
4 & Right subcostal plane (right hepatic lobe, right hepatic vein, diaphragm) & 671 & \textbf{58.72} & 52.61 \\
5 & Right intercostal oblique plane (right hepatic lobe, portal vein, gallbladder) & 682 & \textbf{81.52} & 66.86 \\
6 & Right intercostal oblique plane (right hepatic lobe and right kidney) & 672 & \textbf{85.71} & 76.79 \\
\bottomrule
\end{tabular}}
\end{table}

\paragraph{Additional weak-target summaries.}
On test, full VCRE-Fib has mean inside attention 0.570815, outside response fraction 0.666595, and inside/outside mean-response ratio 359,012.493654; w/o View has 0.499237, 0.550558, and 1,678,411.504391, respectively. The outside response fraction is the summed attention outside the target boxes divided by total attention. The last ratio clips its denominator at $10^{-8}$, so full-field boxes can yield very large values. These are descriptive response summaries, not localization-accuracy measures. Table~\ref{tab:views} compares class recall for the two view-capable models; evaluation artifacts also provide prediction frequencies and confusion matrices.

\subsection{Case selection and display protocol}\label{sec:appendix-case-protocol}
View metrics include accuracy/macro-F1; weak-target Dice/IoU use threshold 0.5. From pools of 12 test and four validation cases, four full-model test cases spanning four true views were selected for the main figure and two validation cases for the appendix. Correspondence candidates were prioritized by weak-box Dice within view. Expansion candidates required an outside response fraction of 0.5--0.9 (summed attention outside the target boxes divided by total attention), Dice above 0.1, a thresholded response intersecting the boxes, and at least 10\% of its area outside all displayed boxes. Patients were deduplicated within each pool. All panels use the same processed ROI, fixed $[0,1]$ localization scale, opacity 0.55, and cyan 0.5 contour. These are selected illustrations; clinical descriptions of the extended regions are not assigned.

\begin{figure}[!htbp]
\centering
\includegraphics[width=\linewidth]{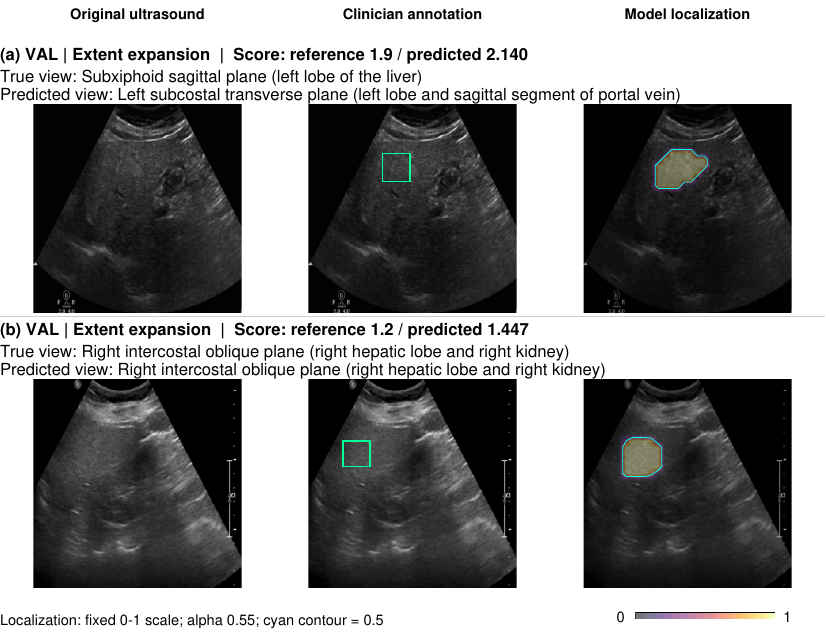}
\caption{Selected validation examples with responses beyond local boxes. Predictions use the frozen checkpoint evaluated quantitatively. The three columns show the original ROI, clinician annotation, and model localization; labels report reference and predicted scores and acquisition views. In the first example, the predicted view differs from the reference. Responses beyond local boxes identify locations for inspection, without establishing complete lesion extent.}
\label{fig:valcases}
\end{figure}

The main paper presents four selected test cases. Figure~\ref{fig:valcases} shows two selected validation examples with responses extending beyond local boxes. These examples are separate from the test ranking and spatial metrics; the first also shows a disagreement between the reference and predicted views.

\FloatBarrier
\section{Validation-selected checkpoints and training diagnostics}\label{sec:appendix-selection}
\begin{figure}[!htbp]
\centering
\includegraphics[width=\linewidth]{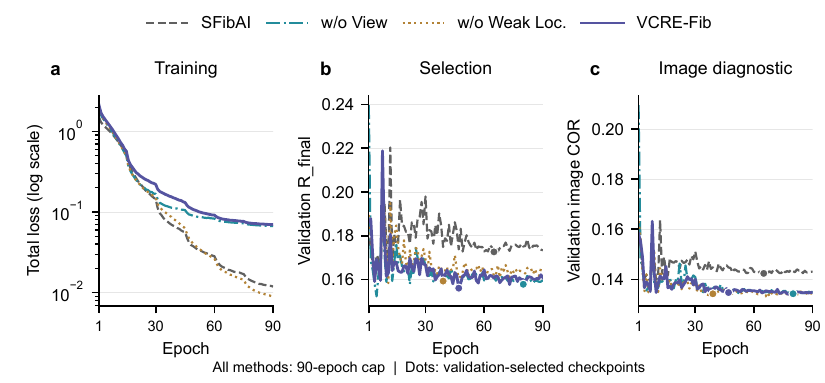}
\caption{\lang{Training and validation trajectories under a common 90-epoch cap. (a) Total training loss on a logarithmic axis, including each model's auxiliary terms. (b) Validation $R_{\mathrm{final}}$. (c) Validation image COR. Curves show unsmoothed epoch-level observations; dots mark the validation-selected checkpoints in Table~\ref{tab:freeze}.}{统一 90 epoch 上限下的训练与验证集轨迹。(a) 含各模型辅助项的训练总损失，纵轴为对数尺度。(b) 验证集 $R_{\mathrm{final}}$。(c) 验证集 image COR。曲线为未经平滑的逐轮观测，圆点标记表~\ref{tab:freeze} 中由验证集选择的检查点。}}\label{fig:training}
\end{figure}

For SFibAI, w/o View, w/o Weak Loc., and full VCRE-Fib, validation selects epochs 65, 80, 39, and 47, with $\risk$ of 0.172661, 0.157844, 0.159315, and 0.156067. Validation contains 20,880 images, 1,227 patients, and 33 centers. Appendix Table~\ref{tab:freeze} records the four frozen checkpoints; Figure~\ref{fig:training} shows their trajectories under the common 90-epoch cap. Test risk determines ranking; validation documents selection.
\begin{table}[!htbp]
\centering\normalsize
\caption{\lang{Validation-selected frozen checkpoints. Validation: 20,880 images, 1,227 patients, 33 centers. All four methods, including VCRE-Fib, use a 90-epoch cap for the training comparison; checkpoints are selected from epochs 21--90. Bold denotes the best unrounded value per metric; exact ties are all bold.}{验证集选择并冻结的检查点；验证集含 20,880 图、1,227 患者、33 中心。包括 VCRE-Fib 在内的四个方法均以 90 epoch 为训练比较上限，检查点从第 21--90 epoch 中选择。各指标按未取整数值加粗最优项，精确并列者均加粗。}}\label{tab:freeze}
\setlength{\tabcolsep}{3pt}
\fitresulttable{%
\begin{tabular}{@{}lrrrr@{}}
\toprule
Method & Epoch cap & Selected epoch & Val $R_{\mathrm{final}}\downarrow$ & Val image COR $\downarrow$ \\
\midrule
SFibAI & 90 & 65 & 0.172661 & 0.142376 \\
w/o View & 90 & 80 & 0.157844 & 0.134258 \\
w/o Weak Loc. & 90 & 39 & 0.159315 & \textbf{0.134234} \\
VCRE-Fib & 90 & 47 & \textbf{0.156067} & 0.134735 \\
\bottomrule
\end{tabular}}
\end{table}

\begin{figure}[!htbp]
\centering
\includegraphics[width=\linewidth]{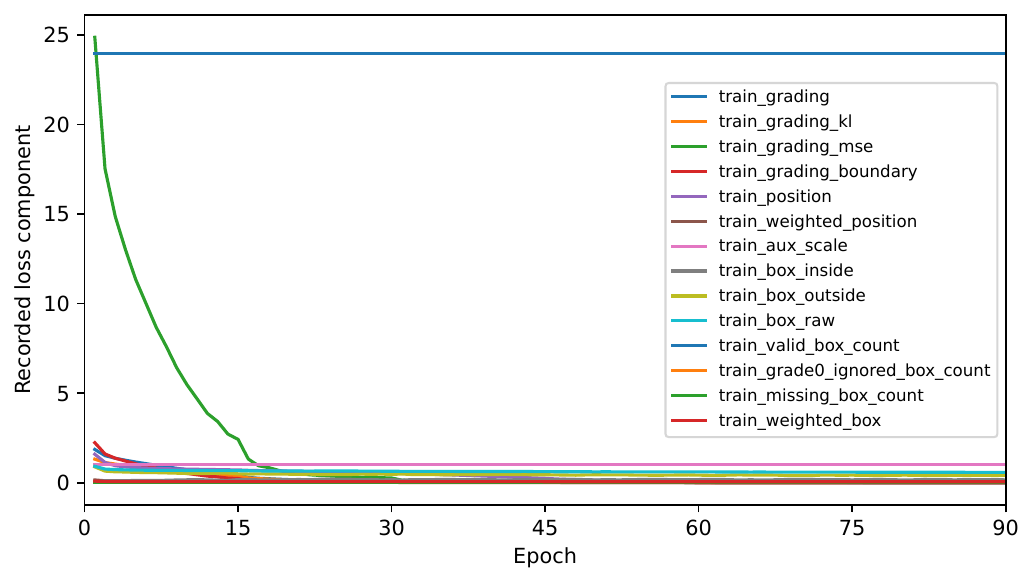}
\caption{\lang{Full VCRE-Fib loss components under the same 90-epoch cap, shown as training diagnostics rather than estimates of module effects.}{完整 VCRE-Fib 在相同 90 epoch 上限下的训练损失分项，用于训练诊断，不作为模块效应估计。}}\label{fig:losses}
\end{figure}

The four models use four GPUs with SyncBatchNorm and global batch 24 under the common training settings in Appendix~\ref{sec:appendix-implementation}. Validation selects from epochs 21--90 by $\risk$, image COR, then earlier epoch; selected checkpoints are frozen before test evaluation. Figure~\ref{fig:training} shows all 90 recorded epochs without smoothing. Total losses contain different auxiliary terms and serve as within-model diagnostics. Figure~\ref{fig:losses} shows the full model's loss components.

\FloatBarrier

\end{document}